%% file: main.tex
\documentclass[runningheads]{llncs}
\usepackage[T1]{fontenc}
\usepackage{graphicx,verbatim}
\usepackage{multirow}
\usepackage[misc]{ifsym}
\usepackage{amsmath}
\usepackage{lipsum}
\usepackage{xstring}
\usepackage{booktabs}
\usepackage{amssymb}

\begin{document}

\title{K-MaT: Knowledge-Anchored Manifold Transport for Cross-Modal Prompt Learning in Medical Imaging}

\titlerunning{K-MaT: Knowledge-Anchored Manifold Transport}
%

\author{Jiajun Zeng\inst{1}
\and
Shadi Albarqouni\inst{1}\textsuperscript{\Letter}
}

\authorrunning{J. Zeng and S. Albarqouni}
%
\institute{University of Bonn, University Hospital Bonn, Clinic for Diagnostic and Interventional Radiology, 53127 Bonn, Germany\\
\email{\{Zeng.Jiajun, Shadi.Albarqouni\}@ukbonn.de}}

\maketitle

\newcommand\blfootnote[1]{%
\begingroup
\renewcommand\thefootnote{}\footnote{#1}%
\addtocounter{footnote}{-1}%
\endgroup
}
\blfootnote{\textsuperscript{\Letter} Corresponding author.}

\begin{abstract}
Large-scale biomedical vision-language models (VLMs) ada\-pted on high-end imaging (e.g., CT) often fail to transfer to frontline low-end modalities (e.g., radiography), collapsing into modality-specific shortcuts. We propose K-MaT (Knowledge-Anchored Manifold Transport), a prompt-learning framework that transfers decision structures to low-end modalities without requiring low-end training images. K-MaT factorizes prompts, anchors them to clinical text descriptions, and aligns the low-end prompt manifold to the visually-grounded high-end space using Fused Gromov-Wasserstein optimal transport. We evaluate K-MaT on four cross-modal benchmarks, including dermoscopy, mammography to ultrasound, and CT to chest X-ray. K-MaT achieves state-of-the-art results, improving the average harmonic mean of accuracy to 44.1\% (from BiomedCoOp's 42.0\%) and macro-F1 to 36.2\%. Notably, on the challenging breast imaging task, it mitigates the catastrophic forgetting seen in standard methods like CoOp (which drops to 27.0\% accuracy on the low-end), preserving robust performance across modalities. Aligning prompt manifolds via optimal transport provides a highly effective route for the zero-shot cross-modal deployment of medical VLMs.


\end{abstract}

\input{sections/intro}
\input{sections/method}
\input{sections/experiments}
\input{sections/conclusions}

\begin{credits}
\subsubsection{\ackname} J. Zeng is funded by the China Scholarship Council (CSC) program (grant No. 202508440222).

\subsubsection{\discintname}
{\small\textbf{The authors have no competing interests.}}
\end{credits}

%
%
%
%

\bibliographystyle{splncs04}
\bibliography{references}
\end{document}

%% file: sections/intro.tex
\section{Introduction}
Deep learning models for medical imaging often degrade under distribution shifts, particularly during cross-modal transfer where distinct acquisition physics encourage modality-specific shortcuts \cite{matta2024systematic,yoon2024domain,gulrajani2020search}. For instance, models trained on high-end screening modalities (e.g., MRI, CT) rarely generalize reliably to accessible, frontline modalities (e.g., X-ray and Ultrasound), despite targeting identical underlying pathologies.

Vision-language models (VLMs) and prompt learning offer efficient adaptation paradigms without full finetuning \cite{clip,biomedclip,coop}. While BiomedCLIP exhibits high generalization, it struggles with specific tasks. CoOp and its variants~\cite{coop,kgcoop} attempt to optimize unified prompts to help models generalize across seen and unseen classes; however, these prompts are prone to overfit to seen classes and domains.
Furthermore, while BiomedCoOp \cite{biomedcoop} improves low-shot biomedical transfer using Large Language Model (LLM)-generated visual descriptions and knowledge distillation from zero-shot BiomedCLIP, it still suffers from similar problems. 
In cross-modal settings, we observe that learnable prompts optimized exclusively on high-end modalities suffer from \textit{catastrophic knowledge forgetting}—they collapse into modality-specific statistics and fail to preserve the essential general textual knowledge required for shared diagnostic semantics~\cite{c-clip}.

To address this, we study asymmetric cross-modal knowledge transfer and ask whether diagnostic semantics learned from high-end visual data can be reliably transferred to low-end modalities in a strict zero-shot regime. We build on the BiomedCLIP~\cite{biomedclip} backbone and propose \textbf{K-MaT (Knowledge-Anchored Manifold Transport)}, a structure-preserving factorized prompt learning framework. Our core contributions are: (i) We introduce a strict zero-shot asymmetric transfer strategy relying solely on high-end visual data and LLM-generated clinical descriptions, eliminating the need for low-end visual training data.
(ii) We mitigate catastrophic forgetting via a novel \textit{space anchoring} constraint, treating LLM-generated textual prototypes as semantic anchors to prevent deviation from clinically meaningful semantics.
(iii) We propose a cross-modal manifold alignment objective using Fused Gromov-Wasserstein (FGW) optimal transport~\cite{titouan2019optimal}, enforcing the learned low-end prompt manifold to strictly mirror the visually grounded relational structure of the high-end manifold.
(iv) Validated on four cross-modal dataset pairs, our approach successfully prevents the decision boundary from collapsing into modality-specific shortcuts, achieving competitive zero-shot generalization against baselines that require target domain data.

%% file: sections/method.tex
\section{Proposed Method}
\begin{figure}[t]
    \centering    \includegraphics[width=\textwidth]{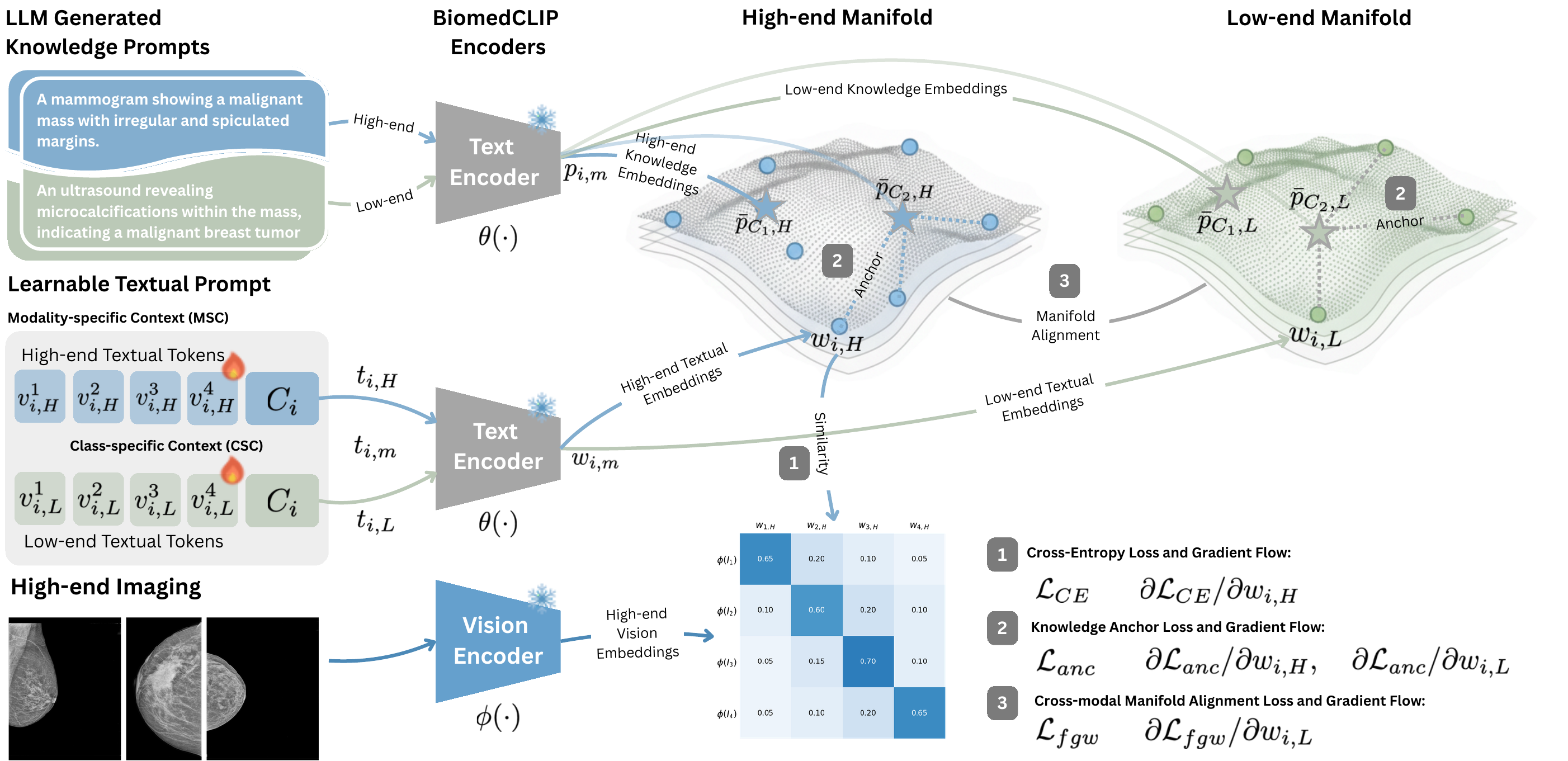}
    \caption{Overall architecture of the proposed K-MaT framework. During training, high-end textual prompts are optimized using high-end imaging data ($\mathcal{L}_{ce}$), while prompts from both modalities are anchored to LLM-generated clinical descriptions ($\mathcal{L}_{anc}$). The low-end prompt manifold is strictly aligned to the visually grounded high-end space via $\mathcal{L}_{fgw}$. During inference, the frozen visual encoder extracts features from unseen low-end images, and predictions are computed via visual-textual similarity with the learned low-end embeddings, bypassing low-end visual training entirely.}
    \label{fig:framework}
\end{figure}

In this section, we detail our proposed K-MaT prompt learning framework for asymmetric cross-modal transfer, as depicted in Fig.~\ref{fig:framework}. Here, comprehensive screening uses a high-end source modality $\mathcal{X}_H$, while frontline assessments rely on a low-end target modality $\mathcal{X}_L$. We aim for High-to-Low generalization by performing few-shot learning on $\mathcal{X}_H$ and zero-shot inference on $\mathcal{X}_L$, entirely bypassing low-end visual data.

\noindent \textbf{Backbone Model and Prompt Parameterization.} 
Following the prompt tuning paradigm, we build upon a frozen biomedical VLM (BiomedCLIP~\cite{biomedclip}) containing a visual encoder $\phi(\cdot)$ and a textual encoder $\theta(\cdot)$. We freeze the pretrained visual and textual encoders and adapt the model via learnable context vectors. For a downstream task consisting of $N_c$ categories, we introduce learnable context vectors to generate task-related textual embeddings. For the $i$-th class token $\mathbf{c}_i$ and modality $m \in \{H, L\}$, we define a factorized learnable prompt:
$$\mathbf{t}_{i,m} = [\mathbf{v}^{1}_{i,m}, \mathbf{v}^{2}_{i,m}, \mathbf{v}^{3}_{i,m}, \mathbf{v}^{4}_{i,m}, \mathbf{c}_{i}]$$
where $\mathbf{v}^{k}_{i,m}$ denote the modality-specific and class-specific learnable context tokens. The specific textual embedding $w_{i,m}$ is obtained by feeding the learnable prompt $\mathbf{t}_{i,m}$ into the textual encoder $\theta(\cdot)$, i.e., $w_{i,m} = \theta(\mathbf{t}_{i,m})$.

\noindent \textbf{High-end Supervised Training.} 
During training, we utilize labeled images exclusively from the high-end modality $\mathcal{X}_H$. Given an image $I_H \in \mathcal{X}_H$ along with its label $y$, the visual embedding is extracted with the visual encoder: $x_H = \phi(I_H)$. 
We optimize the high-end learnable prompts $\mathbf{t}_{i,H}$ by minimizing the standard cross-entropy loss:
$$\mathcal{L}_{ce} = -\frac{1}{N} \sum_{I_H \in \mathcal{X}_H} \log p(y|x_H,w_{i,H})=-\frac{1}{N}\sum_{I_H \in \mathcal{X}_H}{\log \frac{\exp(d(x_H, w_{y,H})/\tau)}{\sum_{i=1}^{N_c} \exp(d(x_H, w_{i,H})/\tau)}}$$
where $\tau$ is a predefined temperature parameter, $N$ is the number of images, and $d(\cdot)$ is the cosine similarity between the visual embedding $x_H$ and the textual embeddings $w_{i,H}$. This objective updates the high-end learnable prompts, explicitly injecting high-end visual knowledge into the textual embedding space.

\noindent \textbf{Space Anchoring to Modality-Specific Lexical Prompts.} 
While learnable prompts capture discriminative task-specific knowledge, they risk collapsing into modality-specific statistics and forgetting the essential general textual knowledge. To alleviate this, we define general textual knowledge using a fixed ensemble of visual descriptions. To reduce the reliance on manual domain expertise while providing precise prior knowledge, we employ an LLM to automatically generate these descriptions. Let $p_{i,m}^j$ denote the $j$-th LLM-generated visual description for the $i$-th class under modality $m$, where $j \in \{1, \dots, N_p\}$ and $N_p$ is the number of descriptions per class. We feed these generated descriptions into the textual encoder to generate the general textual prototype embedding, 
$\bar{p}_{i,m} = \frac{1}{N_p} \sum_{j=1}^{N_p} \theta(p_{i,m}^j)$,
which remains fixed during training. To ensure the learnable specific textual knowledge acts as a semantic anchor and does not deviate from clinically meaningful semantics, we introduce a space anchoring constraint that minimizes the discrepancy (Euclidean distance) between the general textual prototypes and the learnable specific textual embeddings:
$$\mathcal{L}_{anc} = \frac{1}{2N_c} \sum_{i=1}^{N_c} \sum_{m \in \{H,L\}} \| w_{i,m} - \bar{p}_{i,m} \|_2^2$$
As the class and modality tokens are fixed, gradients from $\mathcal{L}_{anc}$ solely update the shared and modality-specific learnable tokens.

\noindent \textbf{Cross-modal Manifold Alignment via FGW.} 
To further prevent the decision boundary from collapsing into high-end specific statistics, we propose a novel alignment objective based on the FGW optimal transport~\cite{titouan2019optimal}. FGW acts as a structural regularizer, enforcing the learned low-end prompt manifold to strictly mirror the relational structure of the high-end prompt manifold anchored by clinical text knowledge.
To explicitly focus this alignment on modality-specific structure, we treat the high-end textual embeddings as a fixed reference manifold. Let $\tilde{w}_{i,H}$ and $\tilde{w}_{j,L}$ denote the textual embeddings for the $i$-th class in the high-end modality and the $j$-th class in the low-end modality, where $i, j \in \{1, \dots, N_c\}$. We first compute the intra-modal structural distance matrices $D^H, D^L \in \mathbb{R}^{N_c \times N_c}$ with elements, 
$D^{H}_{ik} = \|\tilde{w}_{i,H} - \tilde{w}_{k,H}\|_2, \quad D^{L}_{jl} = \|\tilde{w}_{j,L} - \tilde{w}_{l,L}\|_2$,
where indices $i, k$ iterate over the high-end classes and $j, l$ iterate over the low-end classes. Alongside this, we compute the cross-modal feature cost matrix $M \in \mathbb{R}^{N_c \times N_c}$ with elements $M_{ij} = \|\tilde{w}_{i,H} - \tilde{w}_{j,L}\|_2^2$. FGW seeks an optimal coupling matrix $\Gamma \in \mathbb{R}_{+}^{N_c \times N_c}$ to jointly align the feature representations and the relational geometric structure. The corresponding constraint is formulated as:
$$\mathcal{L}_{fgw} = (1 - \alpha) \sum_{i,j=1}^{N_c} M_{ij} \Gamma_{ij} + \alpha \sum_{i,j,k,l=1}^{N_c} |D^{H}_{ik} - D^{L}_{jl}|^2 \Gamma_{ij} \Gamma_{kl}$$
where $\alpha \in [0,1]$ is the trade-off parameter balancing the Wasserstein feature alignment and the Gromov-Wasserstein structural alignment. For FGW, we detach high-end embeddings and backpropagate only into low-end prompt embeddings. Intuitively, $\mathcal{L}_{FGW}$ ensures the low-end prompt manifold preserves the shared diagnostic semantics anchored by the fixed high-end space, entirely bypassing the need for low-end visual training data. 

\noindent \textbf{Overall Objective and Inference.} The final objective combines cross-entropy loss with knowledge anchoring and manifold alignment constraints:
$$\mathcal{L} = \mathcal{L}_{ce} + \lambda_{anc} \mathcal{L}_{anc} + \lambda_{fgw} \mathcal{L}_{fgw}$$
where $\lambda_{anc}$ and $\lambda_{fgw}$ balance the effect of the respective regularization terms. During inference, the learned low-end embeddings $w_{i,L}$ are used to classify unseen images $I_L \in \mathcal{X}_L$ by computing the visual-textual similarity $p(y|x_L, w_{i,L}), \\x_L = \phi(I_L)$ with frozen encoders. During this phase, all visual and textual backbone parameters remain strictly frozen.

%% file: sections/experiments.tex
\section{Experiments and Results}
\subsection{Experimental Settings}

\noindent\textbf{Datasets and Tasks.} We evaluate our proposed K-MaT on 4 diverse cross-modal medical imaging pairs covering various clinical tasks. 
Specifically, the evaluation encompasses: (1) skin lesion classification transferring from Dermoscopy to Clinical Images (DERM7PT)~\cite{derm7pt}; (2) skin lesion classification transferring from Dermoscopy to 15cm Clinical Images (MRA-MIDAS)~\cite{midas}; 
(3) breast lesion classification from Mammography~\cite{dmid} to Ultrasound~\cite{busi}; 
and (4) COVID-19 pneumonia classification from CT~\cite{ccccii} to Chest X-ray~\cite{chestxray}. We also use GPT-5 to generate $N_p=50$ visual descriptions for each class and modality. Table~\ref{tab:datasets} summarizes the dataset statistics and domain characteristics.. 

\noindent\textbf{Implementation Details.}
We adopt the pre-trained BiomedCLIP as our frozen backbone. For prompt tuning, we randomly initialize a learnable context of length 4. To ensure effective transfer and mitigate interference between modality classes, we introduce Class-Specific Context (CSC) and Modality-Specific Context (MSC), learning distinct prompts for each class across different modalities. 

The model is optimized for 50 epochs (batch size 4) using the SGD optimizer with a 0.0025 base learning rate, a cosine annealing scheduler, and a 1-epoch warmup. To compute the FGW optimal coupling $\Gamma$, we use the Sinkhorn-Knopp algorithm~\cite{titouan2019optimal} (temperature 0.1, max 100 iterations) and set the trade-off $\alpha$ to 0.1. All experiments use a 16-shot setting per class, with results averaged over 3 random seeds. Hyperparameters ($\lambda_{anc}$, $\lambda_{fgw}$) are tuned exclusively on the high-end validation set, leaving low-end images strictly for test-time evaluation. We run all experiments on a single NVIDIA A100 (80GB) GPU.
\input{tables/datasets}

\input{tables/results_acc_f1}
\subsection{Experiments on Cross-modal Generalization}
We compare our proposed K-MaT with BiomedCLIP and the state-of-the-art (SOTA) prompt tuning methods equipped with BiomedCLIP backbone, including CoOp~\cite{coop}, CoCoOp~\cite{cocoop}, KgCoOp~\cite{kgcoop}, and BiomedCoOp~\cite{biomedcoop}.Table~\ref{tab:cross_modal_results_acc_f1_combined} details the cross-modal generalization performance in terms of Accuracy and Macro-F1 scores, respectively. Across the four diverse medical image pairs, K-MaT consistently outperforms all competitive baselines, achieving the highest average Harmonic Mean (H) of 44.1\% for Accuracy and 36.2\% for F1 score. H is a critical evaluation metric in our asymmetric setting, as it penalizes extreme performance disparities and accurately reflects the model's ability to retain knowledge in the source high-end modality while effectively generalizing diagnostic semantics to the unseen low-end target modality.

Traditional prompt learning algorithms, such as CoOp and CoCoOp, demonstrate substantial improvements on the high-end source modalities but suffer from severe catastrophic forgetting when evaluated on the low-end target modalities. For instance, in the breast dataset task (transferring from Mammography to Ultrasound), CoOp achieves a high-end accuracy of 75.2\% but plummets to 27.0\% on the Low-end target. Similarly, CoCoOp yields 65.0\% on the High-end but only 28.2\% on the Low-end. These methods tend to overfit to the distinct appearance statistics of the source domain, adopting modality-specific shortcuts rather than shared clinical semantics. K-MaT directly addresses this vulnerability; by preserving structural relationships, it prevents the decision boundary from collapsing, boosting the Low-end target accuracy to 38.4\% on the Breast dataset and achieving a leading H of 50.3\%.

Even when compared to BiomedCoOp, a robust recent baseline tailored specifically for biomedical VLMs, K-MaT exhibits superior generalization. Across all four benchmarks, K-MaT surpasses BiomedCoOp by a 2.1\% margin in average H Accuracy (44.1\% vs. 42.0\%) and a 1.2\% margin in average H F1 (36.2\% vs. 35.0\%). This demonstrates that while integrating general biomedical knowledge is beneficial, it is insufficient on its own to navigate severe cross-modal domain shifts. By explicitly aligning the relational geometry of the prompt manifold through optimal transport, K-MaT effectively distills shared clinical knowledge, successfully achieving competitive zero-shot generalization.

\subsection{Ablation Study}
To evaluate the contribution of each component, we conduct an ablation study to assess the impact of our prompt context designs (CSC and MSC), the space-anchoring loss $\mathcal{L}_{anc}$, and the FGW alignment loss $\mathcal{L}_{fgw}$. Table~\ref{tab:full_ablation_all_configs} presents these architectural variations across High-end, Low-end, and H metrics.

\input{tables/ablation}
\noindent\textbf{Effectiveness of Context Modularization.} The baseline configuration (optimizing solely with cross-entropy $\mathcal{L}_{ce}$) exhibits restricted generalization. It heavily biases predictions toward the high-end, confirming that unconstrained learnable prompts risk collapsing into modality-specific shortcuts. Rows 2 and 3 in Tab.~\ref{tab:full_ablation_all_configs} show that the combination of CSC and MSC can mitigate interference between modality classes and ensure effective cross-modal transfer with the relative improvement from -12.03\% to -5.31\% of Acc. The row 6 and 7 indicate that MSC could improve cross-modal performance (36.67\% Acc) but affect the Acc of high-end. While CSC slightly enhances high-end performance, it disrupts the equilibrium between high-end and low-end modalities.

\noindent \textbf{Effectiveness of Semantic Anchoring and Manifold Alignment.}
Introducing $\mathcal{L}_{anc}$ provides a critical regularizing effect by anchoring learnable embeddings to LLM-generated textual prototypes, preventing semantic divergence and improving the H accuracy to 41.94\% (row 5).  Building on this, $\mathcal{L}_{fgw}$ acts as a powerful structural regularizer. While $\mathcal{L}_{anc}$ maintains semantic fidelity, $\mathcal{L}_{fgw}$ explicitly enforces the low-end prompt manifold to mirror the relational geometry of the high-end space. As shown in Table 3, the complete K-MaT framework—combining CSC, MSC, $\mathcal{L}_{anc}$, and $\mathcal{L}_{fgw}$—achieves the highest overall performance. This full configuration yields a significant 13.75\% relative improvement in H F1 and a 10.10\% boost in H accuracy over the baseline (Row 8).

\begin{figure}[t]
    \centering
    \includegraphics[width=\linewidth]{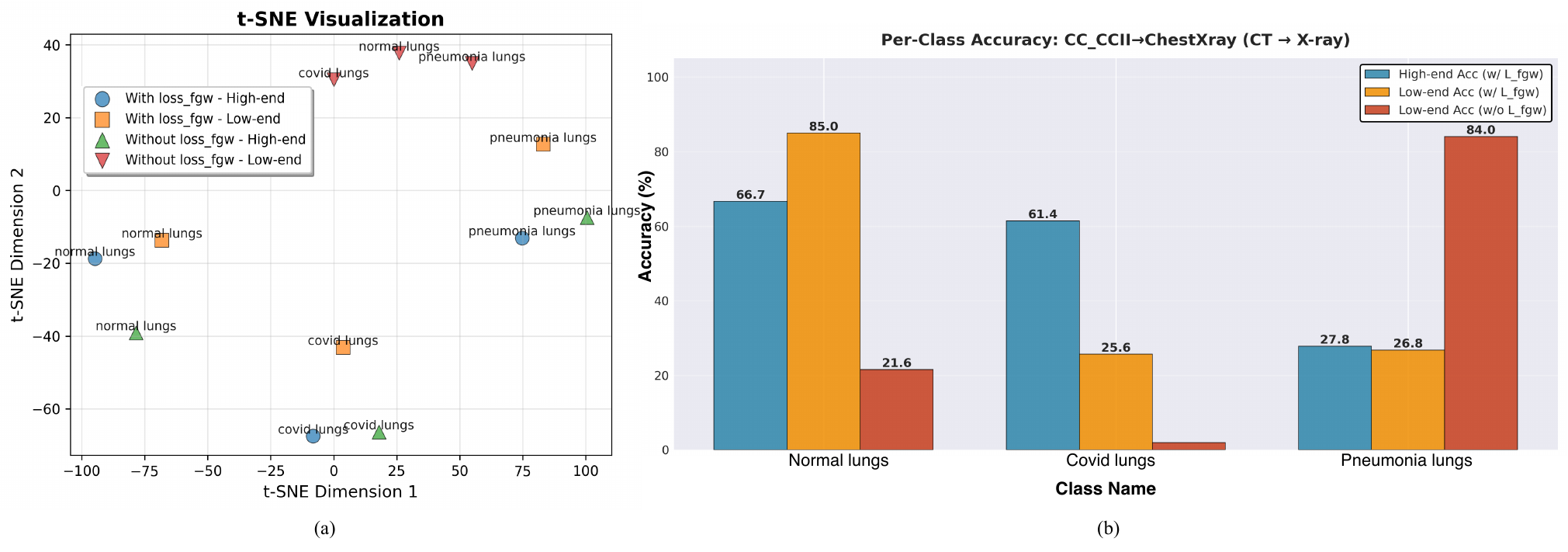}
    \caption{(a). Visualization of textual embeddings w/ and w/o $\mathcal{L}_{fgw}$ on the Chest cross-modal generalization task. (b). The breakdown of Accuracy for each class.}
    \label{fig:discussions}
\end{figure}

\noindent \textbf{Effectiveness of FGW.} To further investigate the role of $\mathcal{L}_{fgw}$, we perform a t-SNE visualization and a breakdown analysis on the chest cross-modal generalization task. Specifically, we compare variants of K-MaT excluding $\mathcal{L}_{anc}$ with and without the addition of $\mathcal{L}_{fgw}$. As observed in Fig.~\ref{fig:discussions}(a), high-end text embeddings $w_{i,H}$ act as anchors due to reliable supervision from labels and visual signals (blue circles and green triangles). By incorporating $\mathcal{L}_{fgw}$, the low-end embeddings $w_{i,L}$ tend to preserve the relative structure of their high-end counterparts (yellow squares). In contrast, text embeddings learned without $\mathcal{L}_{fgw}$ fail to generate discriminative representations (red triangles). This is further confirmed by the breakdown analysis in Fig.~\ref{fig:discussions}(b): $\mathcal{L}_{fgw}$ prevents the model from collapsing into a single class, improving average performance on the low-end modality through structural transport from the high-end space.

%% file: tables/datasets.tex
\begin{table}[t!]
\centering
\fontsize{8pt}{\baselineskip}\selectfont
\caption{Dataset statistics for the five cross-modal tasks.}
\label{tab:datasets}
\begin{tabular}{lllllccc}
\toprule
\textbf{Task} & \textbf{Type} & \textbf{Dataset} & \textbf{Modality} & \textbf{Classes} & \textbf{Train} & \textbf{Val} & \textbf{Test} \\
\midrule
\multirow{2}{*}{Skin I} & High-end & MRA-MIDAS~\cite{midas} & Dermoscopy & 6 & 513 & 91 & 135 \\
 & Low-end & MRA-MIDAS~\cite{midas} & Clinical & 6 & 501 & 89 & 132 \\
\midrule
\multirow{2}{*}{Skin II} & High-end & Derm7pt~\cite{derm7pt} & Dermoscopy & 5 & 413 & 203 & 395 \\
 & Low-end & Derm7pt~\cite{derm7pt} & Clinical & 5 & 413 & 203 & 395 \\
\midrule
\multirow{2}{*}{Chest} & High-end & CC-CCII~\cite{ccccii}& CT & 3 & 10670 & 3538 & 1678 \\
 & Low-end & ChestXray~\cite{chestxray} & X-Ray & 3 & 4595 & 923 & 914 \\
\midrule
\multirow{2}{*}{Breast} & High-end & DMID~\cite{dmid} & Mammography& 3 & 356 & 76 & 78 \\
 & Low-end & BUSI~\cite{busi} & Ultrasound & 3 & 389 & 155 & 236 \\
\bottomrule
\end{tabular}
\end{table}

%% file: tables/results_acc_f1.tex
\begin{table*}[t]
    \centering
    \caption{Cross-modal generalization performance on 4 medical image pairs with ACC and macro-F1 shown together. Each entry is reported as average ACC (F1) over 3 seeds. Hi: High-End (source), Lo: Low-End (target), H: Harmonic Mean. Best ACC/F1 results are highlighted in \textbf{bold}/\underline{underline} respectively.}\label{tab:cross_modal_results_acc_f1_combined}
\fontsize{8pt}{\baselineskip}\selectfont
\begin{tabular}{cccccccc}
\toprule
\multicolumn{2}{c}{\textbf{Dataset}} & BiomedCLIP & CoOp & KgCoOp & CoCoOp & BiomedCoOp & K-MaT \\
\midrule
\multirow{3}{*}{Skin I}
& Hi & 34.6 (24.2) & 55.5 (\underline{42.3}) & 56.6 (41.6) & 44.6 (32.0) & \textbf{58.3} (39.3) & \underline{56.8} (\textbf{43.1}) \\
& Lo & 31.7 (21.0) & 40.8 (27.1) & 40.5 (\underline{28.5}) & 29.8 (24.2) & \underline{42.2} (\textbf{29.1}) & \textbf{46.4} (27.4) \\
& H  & 37.9 (22.9) & 47.0 (33.0) & 47.2 (\textbf{33.9}) & 35.7 (27.5) & \underline{49.0} (\underline{33.4}) & \textbf{50.9} (33.3) \\
\midrule
\multirow{3}{*}{Skin II}
& Hi & 32.6 (26.9) & 34.8 (32.1) & 34.6 (32.7) & \textbf{40.7} (\textbf{36.7}) & \underline{35.6} (31.8) & 34.8 (\underline{32.8}) \\
& Lo & 17.4 (13.6) & \underline{28.5} (\underline{26.4}) & \underline{28.5} (\textbf{27.3}) & 18.4 (18.0) & 27.0 (24.6) & \textbf{31.6} (25.0) \\
& H  & 22.7 (18.1) & \underline{31.4} (\underline{29.0}) & 31.3 (\textbf{29.8}) & 25.4 (24.2) & 30.7 (27.7) & \textbf{33.1} (28.3) \\
\midrule
\multirow{3}{*}{Chest}
& Hi & 34.0 (24.6) & \underline{52.0} (\underline{47.8}) & 47.1 (42.3) & 37.4 (36.3) & 43.4 (40.9) & \textbf{54.5} (\textbf{50.8}) \\
& Lo & 29.9 (\underline{32.1}) & 26.4 (21.8) & 26.1 (22.7) & \textbf{54.5} (\textbf{35.6}) & 38.7 (29.2) & \underline{40.6} (31.1) \\
& H  & 31.8 (27.9) & 35.0 (30.0) & 33.6 (29.5) & \textbf{44.4} (\textbf{35.9}) & 40.9 (34.0) & \underline{42.3} (\underline{35.4}) \\
\midrule
\multirow{3}{*}{Breast}
& Hi & 24.4 (19.9) & \textbf{75.2} (\textbf{72.8}) & 70.5 (66.6) & 65.0 (59.5) & 69.2 (65.0) & \underline{73.1} (\underline{70.7}) \\
& Lo & \textbf{57.2} (\textbf{45.2}) & 27.0 (19.0) & 28.0 (16.9) & 28.3 (26.6) & 36.0 (34.1) & \underline{38.4} (\underline{36.3}) \\
& H  & 34.2 (27.7) & 39.7 (30.2) & 40.1 (27.0) & 39.4 (36.8) & \underline{47.4} (\underline{44.8}) & \textbf{50.3} (\textbf{47.9}) \\
\midrule
\multirow{3}{*}{\textbf{Avg.}}
& Hi & 34.6 (24.2) & \underline{54.4} (\underline{48.7}) & 52.2 (45.8) & 46.9 (41.1) & 51.6 (44.2) & \textbf{54.8} (\textbf{49.3}) \\
& Lo & 34.0 (28.0) & 30.7 (23.6) & 30.8 (23.9) & 32.7 (26.1) & \underline{36.0} (\underline{29.2}) & \textbf{39.3} (\textbf{29.9}) \\
& H  & 31.7 (24.1) & 38.3 (30.5) & 38.0 (30.0) & 36.2 (31.1) & \underline{42.0} (\underline{35.0}) & \textbf{44.1} (\textbf{36.2}) \\
\bottomrule
\end{tabular}
\end{table*}

%% file: tables/ablation.tex
\begin{table*}[t]
    \centering
    \fontsize{8pt}{\baselineskip}\selectfont
    \caption{Complete ablation study results across all dataset pairs. Relative improvement (\%) is computed on the Harmonic Mean ACC/F1 with respect to the first row.}
    \label{tab:full_ablation_all_configs}
    \begin{tabular}{cccc cccccc cc}
    \toprule
        \multirow{2}{*}{\textbf{CSC}} & \multirow{2}{*}{\textbf{MSC}} & \multirow{2}{*}{\textbf{$\mathcal{L}_{\text{anc}}$}} & \multirow{2}{*}{\textbf{$\mathcal{L}_{\text{fgw}}$}} 
    & \multicolumn{2}{c}{\textbf{High-end}} 
    & \multicolumn{2}{c}{\textbf{Low-end}} 
    & \multicolumn{2}{c}{\textbf{Harmonic Mean}}
    & \multicolumn{2}{c}{\textbf{Rel. Impr. (\%)}} \\
    \cmidrule(lr){5-6} \cmidrule(lr){7-8} \cmidrule(lr){9-10} \cmidrule(lr){11-12}
    & & & & \textbf{ACC} & \textbf{F1} & \textbf{ACC} & \textbf{F1} & \textbf{ACC} & \textbf{F1} & \textbf{ACC} & \textbf{F1} \\
    \midrule
     &  &  &  & 53.36 & 47.53 & 35.27 & 25.72 & 40.08 & 31.85 & 0.00 & 0.00 \\
    \checkmark &  &  &  & 54.02 & 48.49 & 27.55 & 22.71 & 35.26 & 30.03 & -12.03 & -5.71 \\
    \checkmark & \checkmark &  &  & 54.69 & 49.17 & 34.73 & 18.68 & 37.95 & 25.83 & -5.31 & -18.90 \\
    \checkmark & \checkmark & \checkmark &  & 54.80 & 49.34 & 30.95 & 20.73 & 35.15 & 27.82 & -12.30 & -12.65 \\
    \checkmark & \checkmark &  & \checkmark & 54.69 & 49.17 & 36.99 & 26.58 & 41.94 & 32.65 & 4.64 & 2.51 \\
     & \checkmark & \checkmark & \checkmark & 50.68 & 44.70 & 30.07 & 24.24 & 36.67 & 30.67 & -8.51 & -3.70 \\
    \checkmark &  & \checkmark & \checkmark & 53.56 & 48.07 & 28.18 & 22.94 & 35.92 & 30.29 & -10.38 & -4.90 \\
    \midrule
    \checkmark & \checkmark & \checkmark & \checkmark & \textbf{54.80} & \textbf{49.34} & \textbf{39.25} & \textbf{29.92} & \textbf{44.13} & \textbf{36.23} & \textbf{10.10} & \textbf{13.75} \\
    \bottomrule
    \end{tabular}
\end{table*}

%% file: sections/conclusions.tex
\section{Conclusion}
We proposed K-MaT, a zero-shot prompt learning framework for asymmetric cross-modal transfer in medical imaging. By factorizing prompts, anchoring them to LLM-generated clinical text, and aligning manifolds via FGW optimal transport, K-MaT effectively mitigates catastrophic forgetting and achieves SOTA harmonic mean performance across four benchmarks. Despite these promising results, our framework exhibits certain limitations. First, its absolute performance on low-end modalities shows limited improvement over the zero-shot BiomedCLIP baseline. Second, the framework is sensitive to dataset characteristics; severe visual discrepancies between modalities can create visual-textual gaps that purely text-anchored alignment cannot fully bridge. 
Future work will explore incorporating more reliable visual signals to enhance the stability and low-end transfer capabilities of the framework.